\documentclass[conference]{IEEEtran}
\IEEEoverridecommandlockouts
\usepackage{cite}
\usepackage{amsmath,amssymb,amsfonts}
\usepackage{algorithmic}
\usepackage{graphicx}
\usepackage{textcomp}
\usepackage{xcolor}
\def\BibTeX{{\rm B\kern-.05em{\sc i\kern-.025em b}\kern-.08em
    T\kern-.1667em\lower.7ex\hbox{E}\kern-.125emX}}

\usepackage{subfig}
\usepackage{algorithm2e}
\usepackage{cleveref}

\begin{document}
\title{Optimizing  Convergence for Iterative Learning of ARIMA for Stationary Time Series
}
\author{
\IEEEauthorblockN{Kevin Styp-Rekowski}
\IEEEauthorblockA{Distributed and Operating Systems\\
TU Berlin\\
Berlin, Germany\\
styp-rekowski@tu-berlin.de}
\and
\IEEEauthorblockN{Florian Schmidt}
\IEEEauthorblockA{Distributed and Operating Systems\\
TU Berlin\\
Berlin, Germany\\
florian.schmidt@tu-berlin.de}
\and
\IEEEauthorblockN{Odej Kao}
\IEEEauthorblockA{Distributed and Operating Systems\\
TU Berlin\\
Berlin, Germany\\
odej.kao@tu-berlin.de}
}

\maketitle

\begin{abstract}
Forecasting of time series in continuous systems becomes an increasingly relevant task due to recent developments in IoT and 5G. The popular forecasting model ARIMA is applied to a large variety of applications for decades. An online variant of ARIMA applies the Online Newton Step in order to learn the underlying process of the time series. This optimization method has pitfalls concerning the computational complexity and convergence. Thus, this work focuses on the computational less expensive Online Gradient Descent optimization method, which became popular for learning of neural networks in recent years.
For the iterative training of such models, we propose a new approach combining different Online Gradient Descent learners (such as Adam, AMSGrad, Adagrad, Nesterov) to achieve fast convergence. The evaluation on synthetic data and experimental datasets show that the proposed approach outperforms the existing methods resulting in an overall lower prediction error. 
\end{abstract}


\section{Introduction}
\label{introduction}

In recent years, the advances in technology enable the broad collection of data of e.g. continuously streams provided by the Internet of Things (IoT), Industry 4.0, Social media, etc. 
Due to these rapid developments, big data processing and especially stream processing became relevant. A prominent task is the forecasting of the stream signal to e.g. foresee trends, future abnormal behavior in order to benefit from early responses like in case of predictive maintenance. 

In context of forecasting, the algorithm ARIMA \cite{box2015time} is widely applied including in financial forecasts \cite{raymond1997application}, energy forecasts \cite{sen2016application}, and many other applications with underlying cyclic time series patterns \cite{yu2016improved, schmidt2018unsupervised1}. 
Past observations are used to predict the future behaviour by learning or imitating the underlying process which generates the time series data.
ARIMA assumes that the observed time series relies on a stationary process learned by the model. ARIMA models are typically trained statically on a fixed set of data points. 
In the context of data streams, the time series may underlay concept drifts and consequently continuous online training provides the possibility to adapt to the current time series behavior. Liu et. al \cite{liu2016online} proposed Online ARIMA, which provides optimized iterative training to the traditional ARIMA model by applying Online Gradient Descent (OGD) convex optimization. 

For any forecasting task the accuracy must be at highest possible level which means that the error between the forecasted and observed values shall be minimal (residuals). 
Liu et. al \cite{liu2016online} did not investigate in detail variants of OGD to achieve low residuals as quick as possible, while also not ending up in a local optimum in order to achieve at the long run optimized low residuals.



OGD convex optimization solver enables this learning in an online fashion.
There are several optimizers for the OGD approach which enable faster learning with the OGD optimization solver. Previous research has shown that these optimizers have pitfalls \cite{reddi2019convergence, tran2019convergence,chen2018convergence} in terms of their convergence properties, meaning that they sometimes diverge and sometimes do not converge to the optimal solution.

Therefore, this work focuses on extending the learning capabilities of Online ARIMA.
We propose an optimization strategy by combining different Online Gradient Descent optimizers to achieve a high initial training progress by applying adaptive methods at the beginning and gradually switching to non-adaptive methods to benefit from their generalization ability, resulting in fine-tuned coefficients for the underlying process.
Our paper makes the following key contributions:
\begin{itemize}
    \item Design and development of a combined optimizer for learning Online ARIMA models.
    \item Experimental evaluation of the proposed approach and its hyperparameters using representative synthetic and real-world IoT-sensor datasets.
\end{itemize}

The rest of the paper is structured as follows: Section II introduces Online ARIMA and further background. Section III provides the related work concerning optimizers. Afterwards, the proposed approach of combining the optimizers of the Online ARIMA model during training is described in Section IV, which is evaluated and discussed in Section V. Finally, Section VI concludes the paper. 

\section{Background}

\subsection{Online ARIMA}
Liu et. al \cite{liu2016online} proposed Online ARIMA (AutoRegressive Integrated Moving Average) by approximating the original ARIMA(k,d,q) model with an Online ARIMA(mk,d,0) model, whilst giving estimates on the regret bound while learning with the proposed online convex optimization solver Online Gradient Descent (OGD). In order to learn the coefficients of the Online ARIMA model these optimization solvers run in a continuous system and optimize the coefficients of the ARIMA model to forecast the time series with a smaller error, thus reflecting the time series more precisely. The Online ARIMA model uses following formula to forecast the next point $\widetilde{X}_t$:

\begin{equation}
    \widetilde{X}_t = \sum_{i=1}^{mk} \gamma_i \nabla^d X_{t-i} + \sum_{i=0}^{d-1} \nabla^i X_{t-1}
\end{equation}

The past seen values of $X$ are multiplied with the coefficient vector $\gamma$ with respect to the grade of derivation defined by $d$. The residual value, also known as the error or the distance during prediction, is defined as the distance between the predicted forecasted value $\widetilde{X}_t$ and the real value $X_t$. The lower the residual value of a forecasting model, the better the quality is considered. The model learns the coefficients which weights the seen points and calculate a prediction for the next point.

\subsection{OGD Optimizers}

The OGD optimization solver is widely used and has a lot of versions which vary the gradient updates that are applied to the coefficients \cite{zinkevich2003online}. The OGD optimization solver's computational complexity grows linear with the input size. Thus, this work will focus on the computational less expensive OGD optimizer. The basic implementation considers the gradient of the loss function $\nabla l(\theta)$ and updates the coefficients $\theta$ based on the negative gradient by a certain factor, the learning rate $\eta$ at time $t$:

\begin{equation}
    \theta_t = \theta_{t-1} - \eta \nabla l(\theta_t)
\end{equation}

More sophisticated optimizers like the Momentum \cite{qian1999momentum} or Nesterov \cite{nesterov1983method} optimizer take into account the momentum, in terms of direction of the changes in the last iterations by adding a term representing the changes of the last iterations.

Another class of adaptive optimizers have evolved including the methods AdaGrad \cite{duchi2011adaptive}, RMSProp \cite{tieleman2012lecture}, Adam \cite{kingma2014adam} and AMSGrad \cite{reddi2019convergence}. These optimizers take into account several factors during the optimization process and use individual adaptive learning rates for each parameter. Adam uses the momentum, as well as the previous update as well as factors correcting both calculated momentum values. Adagrad, RMSProp and AMSGrad are  similar to Adam with certain differences within the applications of the gradient to the upgrade of the coefficients.

\section{Related Work}
Besides convex optimization solvers, there also exist evolutional greedy approaches as genetic algorithms or simulated annealing \cite{back1996evolutionary}. These algorithms try to find optima by exploring the search space and exploiting enhancements to already found solutions. Due to their greedy property these algorithms might not approach the global optimum and cannot guarantee a quality increase  of the model during learning. Thus, convex optimization solvers are used.

Related work  improves the convergence of the OGD optimizer by weighting the updates based on the importance of the samples and thus achieving faster convergence \cite{karampatziakis2010online}. This technique is also referred to as importance sampling and yields benefits for the convergence bounds \cite{needell2016stochastic}. However, these techniques are not feasible for an online setting, as the importance of samples cannot be obtained in a universally valid way.

Research investigates the combination of different learning rates and their adaptation, called learning rate schedule \cite{ruder2016overview, darken1992learning}. Several methods exploit  pre-defined schedules, which also have their pitfalls in terms of tuning the parameters of the schedule. Another possibility is to adapt the learning rate automatically depending  on the gradient, which is able to adapt fast to changes but also in fine-tuning the model \cite{schaul2013no}. This fast adaptation to changes is unfeasible if outliers are present. Recent work also proposed the learning of the learning rate itself \cite{ravaut2018gradient} with dedicated models. In their work they only consider the basic OGD optimizer.
To our best of knowledge, there is no related work investigating the combination of different OGD optimizers for ARIMA models.

For the training of neural networks, similar approaches of combining optimizers were applied. SWATS \cite{keskar2017improving} is an approach which switches from Adam to SGD at a certain point during training. This switching point is critical for the optimization process and as the authors warn, switching too early does not yield the benefits from Adam's initial progress while switching too late does not yield any generalization improvements. Therefore, Luo et al. \cite{luo2019adaptive} propose a smooth conversion of optimizers with AdaBound. AdaBound clips the learning rates of Adam and AMSGrad with lower and upper bounds, which gradually restrict the learning rate to eventually evolve to the same lower and upper bound, resulting in the SGD optimizer.  

\section{Combination of Optimizers}
The Online ARIMA model shall represent the underlying process of the time series as fast as possible in a continuous system. There is a short initial set-up phase as ARIMA needs to collect at least $mk + d$ samples in order to fill its history of seen data points and gradients with meaningful data. Only from this point on the forecasted values can be used for evaluating the model and further learning of the coefficients.


\subsection{General Idea}

A trade-off between adaptive and non-adaptive methods has to be considered. Adaptive optimizers converge efficiently to a good set of coefficients but oftentimes fail to find the best solution. Non-adaptive methods have the ability to learn a hidden process precisely but are unable to adapt the learning rates for a faster convergence. The proposed approach aims at a combination that reduces the shortcomings of the approaches. Thus, we propose a combination of different OGD optimizers in order to converge fast to a near-optimal solution and gradually switching to a non-adaptive learner to learn the underlying process as precise as possible. This also enables the model to adapt to small concept shifts in the underlying process while making it robust to outliers.


We propose using AMSGrad as the adaptive method and the momentum OGD implementation as the non-adaptive method. The AMSGrad estimates the model of the underlying process quite well while the momentum OGD optimizer is able to optimize the model further, at the same time being able to adapt to concept shifts and being resistant to abnormal changes like outliers.

\subsection{Combined Optimizer}

For the combination of the optimizers we propose a linearly graduating combination of the optimizers $h_t$. The resulting gradients of the AMSGrad and Momentum optimizers are calculated separately and then combined, based on our linear function. Based on the current timestep $t$ the resulting combination of the calculated gradients is used as follows. With gradient at current timestep $g_t$, current AMSGrad gradient $a := AMSGrad(g_t)$, current Momentum gradient $m := Momentum(g_t)$ and hyperparameter $\lambda$:
\begin{equation}
    h_t = \Bigg\{
        \begin{array}{ll}
        (1 - \frac{t}{\lambda}) * a + \frac{t}{\lambda} * m & \text{, } if \text{ } 0\leq \frac{t}{\lambda}\leq 1\\
        m & \text{, } else \text{ (} \frac{t}{\lambda} > 1  \text{)} 
        \end{array}
\end{equation}

By gradually applying both optimizers during the learning process of an Online ARIMA model an optimization to the overall convergence process can be achieved. The adaptive AMSGrad optimizer has high learning speed at the beginning while the non-adaptive momentum optimizer guarantees a precisely learned model while being able to adapt to concept changes. This is a reduction to the overall residual during prediction over several iterations. The evaluation will show that the introduced hyperparameter $\lambda$ is quite insensitive and even edge configurations lead to an improved overall residual.

\section{Evaluation}
\setlength{\columnsep}{0.2 in}
\def\BibTeX{{\rm B\kern-.05em{\sc i\kern-.025em b}\kern-.08em T\kern-.1667em\lower.7ex\hbox{E}\kern-.125emX}}

As ARIMA is a model for the forecasting of time series the timely order of the samples is important. The proposed method will be evaluated on synthetic and real world data on Online ARIMA models.

\subsection{Synthetic Data}

The evaluation contains three settings for the synthetic data where each setting consists of 10,000 samples. The data is generated using different ARIMA models, thus varying the task of online learning data. In the online setting, for each incoming sample one step of the convex optimization solvers is executed. The data is normalized in the interval of $[-1, 1]$.

The first setting is an ARIMA model with the parameters $d=0$, $\alpha = [0.9, -0.9, 0.9, -0.4, -0.1]$, $\beta = []$. This describes a stationary process with no noise as the $\beta$-Terms are left out. The second setting utilizes similar parameters as the first setting with the addition of noise within the Moving Average-Terms, manifested in the $\beta$-coefficients: $[0.5, 0.1]$.

The third setting uses the same ARIMA model as in the second setting for the first 5,000 samples. The second 5,000 samples of this dataset shall simulate a concept shift within the data. Therefore, the ARIMA model used for the second half of the dataset uses the parameters $d=0$, $\alpha_2 = [0.7, -0.7, 0.7, -0.6, -0.3]$, $\beta_2 = [0.2,0.4]$. We defined the parameters $\alpha$, $\beta$, $\alpha_2$ and $\beta_2$ to generate a stationary process which expresses itself in different observed functions switching from $\alpha$, $\beta$ configuration to $\alpha_2$, $\beta_2$.

\subsection{Real-world High-frequency Datasets}

For the real world datasets, the learning will be based on micro-batches resulting from a tumbling window. This means that the shown results are averaged over these micro-batches where at each sample of the micro-batch one step of the convex optimization solvers is executed and a change is applied to the coefficients of the Online ARIMA model.

In the used datasets there are natural occurences of micro-batches which are used similarly during the evaluation as this reflects the real world setting best. The data consists of measurements of the real world which are taken by sensors measuring at described intervals one mini-batch of samples.

\subsubsection{NASA Bearing Dataset}

The bearing dataset provided by Lee et al. \cite{lee2007bearing} contains measured vibration data from accelerometers attached to the bearing housing attached to a rotating shaft, simulating a rotating machine. In particular, the second test executed contains data for one axis of four bearings, resulting in 4 measured channels of vibration data. At the end of this test, there occurs an outer race failure at the first bearing. This describes one of the use cases where learning the underlying process as fast and precisely as possible is very important as this enables the possibility to detect divergent behaviour compared to the forecasted values. The data was measured with a sampling rate of 20kHz and 20,480 samples every 10 minutes, resulting in micro-batches of 20,480 samples. For our experiments only the data of the first bearing was used.

\subsubsection{Industry dataset}
A second industry dataset contains similar 3-axis (x, y, z) vibration data which was measured by acceleration sensors at the machine housing of a rotating machine. The  data  was  measured  with  a  sampling rate of 2kHz and 10,240 samples every hour, resulting in micro-batches of 10,240 samples. For this dataset one axis was chosen to exemplary evaluate the approach.

As the vibration is physically bound to the machine, both datasets represent stationary processes and systems which operate in a continuous setting. A precise forecast of such systems can be beneficial in a variety of use cases, e.g. anomaly detection. Abnormal behavior of the time series can be detected by a abnormal difference between the real values and the forecasted values, thus emphasizing a fast adaptation and low residual values.

\subsection{Evaluation Metric}

For measuring the error of the Online ARIMA models, we define the average residual value $r_t$ as the evaluation metric. The average residual value $r_t$ is the inherent ARIMA error resulting from the difference of the predicted value $\widetilde{x_{i,t}}$ to the real value $x_{i,t}$ at each point $i$ within the specified micro-batch $t$ or sample $t$ of the time series, respectively. For the micro-batch setting, this residual value is calculated by averaging the absolute error for each predicted sample with the size of the micro-batch $n$, representing the mean absolute error:

\begin{equation}
    r_t= \sum_{i=mk+d}^{n} \text{ } \lower3px\hbox{$\dfrac{ | \widetilde{x_{i,t}} - x_{i,t} |}
    {n - mk - d}$}
\end{equation}

The initial $mk+d$ samples, needed to set up the ARIMA model for predicting, are not considered. This average distance is a good measurement for the quality of the ARIMA model, as it measures how precise the model predicts with low error.


For a comparison of the proposed approach against the existing approaches, the same data is iterated over multiple times using different optimizers. As the Online ARIMA models are randomly initialized, the experiments are repeated several times and the averaged results are shown.

As stated previously, tuning the learning rate is not in the scope of this paper, therefore the same fixed learning rate is used for the different approaches. Hereby, the learning rates were chosen using grid search with varying step sizes and learning rates in the range of $[0, 1]$.

\subsection{Results}

\begin{figure}[t]
\centering
    \includegraphics[width=3.05in]{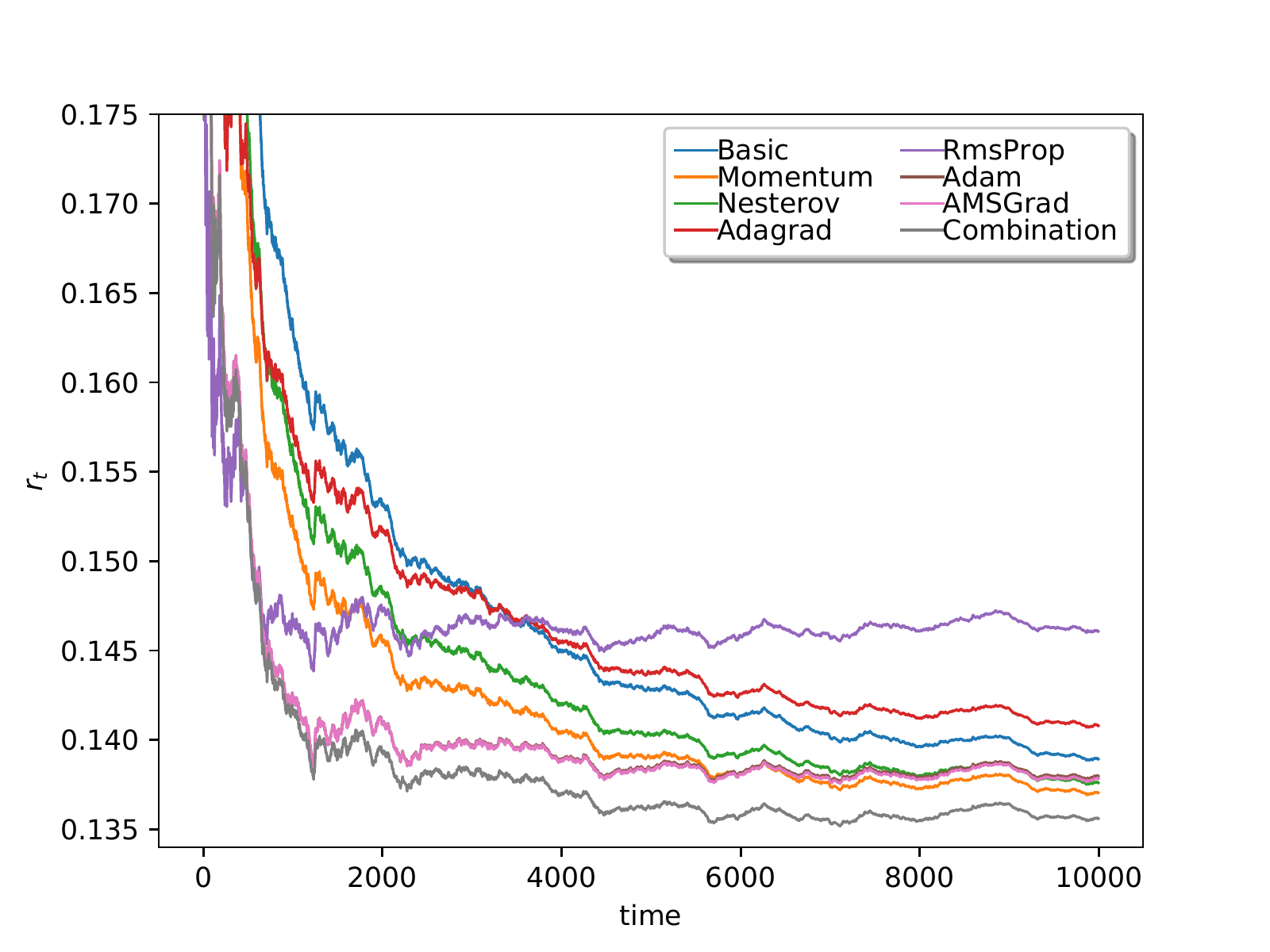}
    \caption{Average residual value for ARIMA(5,0,0) data.}
    \label{fig:arima_5_0}
\end{figure}

\begin{figure}[t]
    \centering
    \includegraphics[width=3.05in]{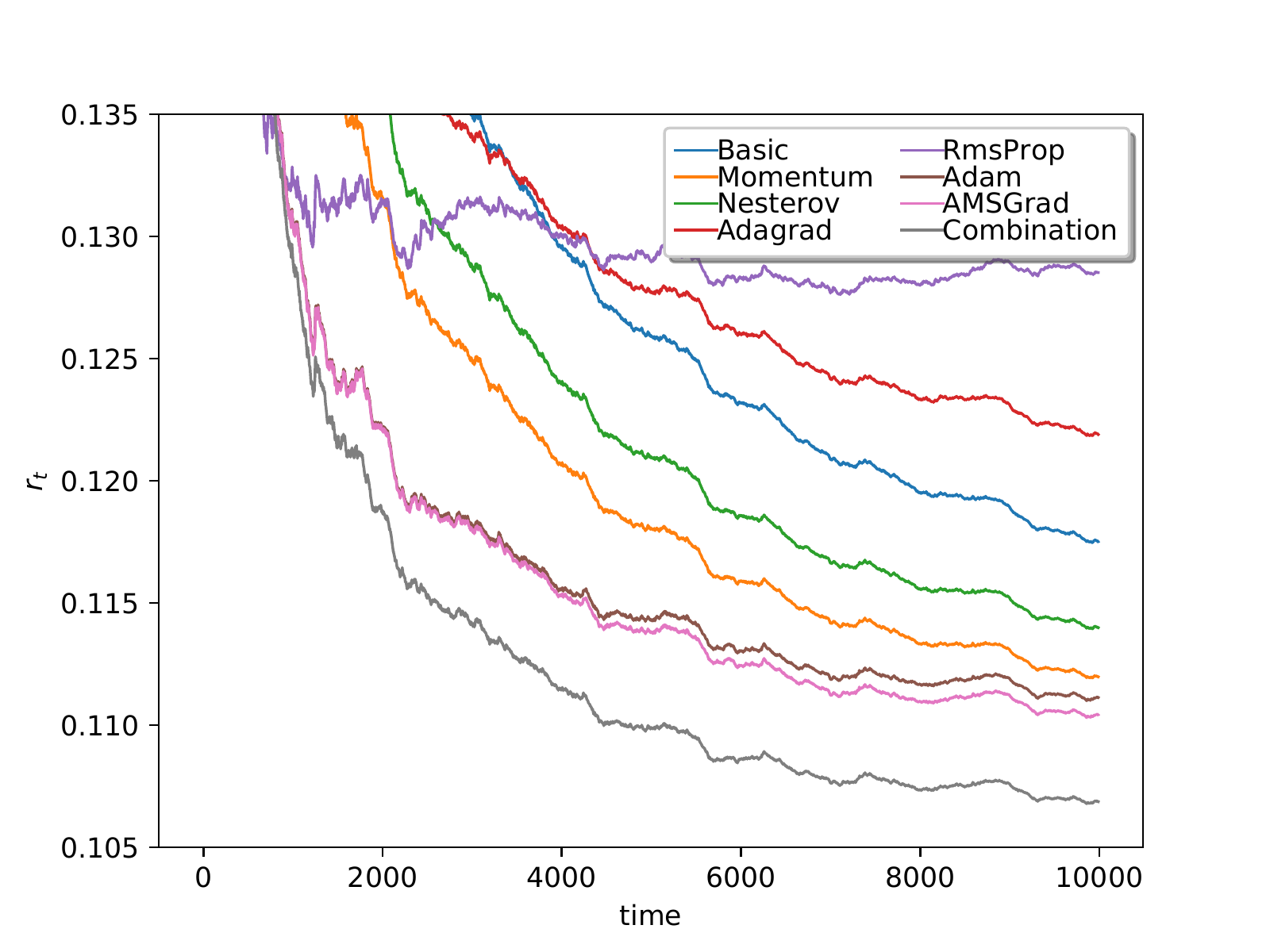}
    \caption{Average residual value for ARIMA(5,0,2) data.}
    \label{fig:arima_5_2}
\end{figure}

\begin{figure}[t]
\centering
    \includegraphics[width=3.05in]{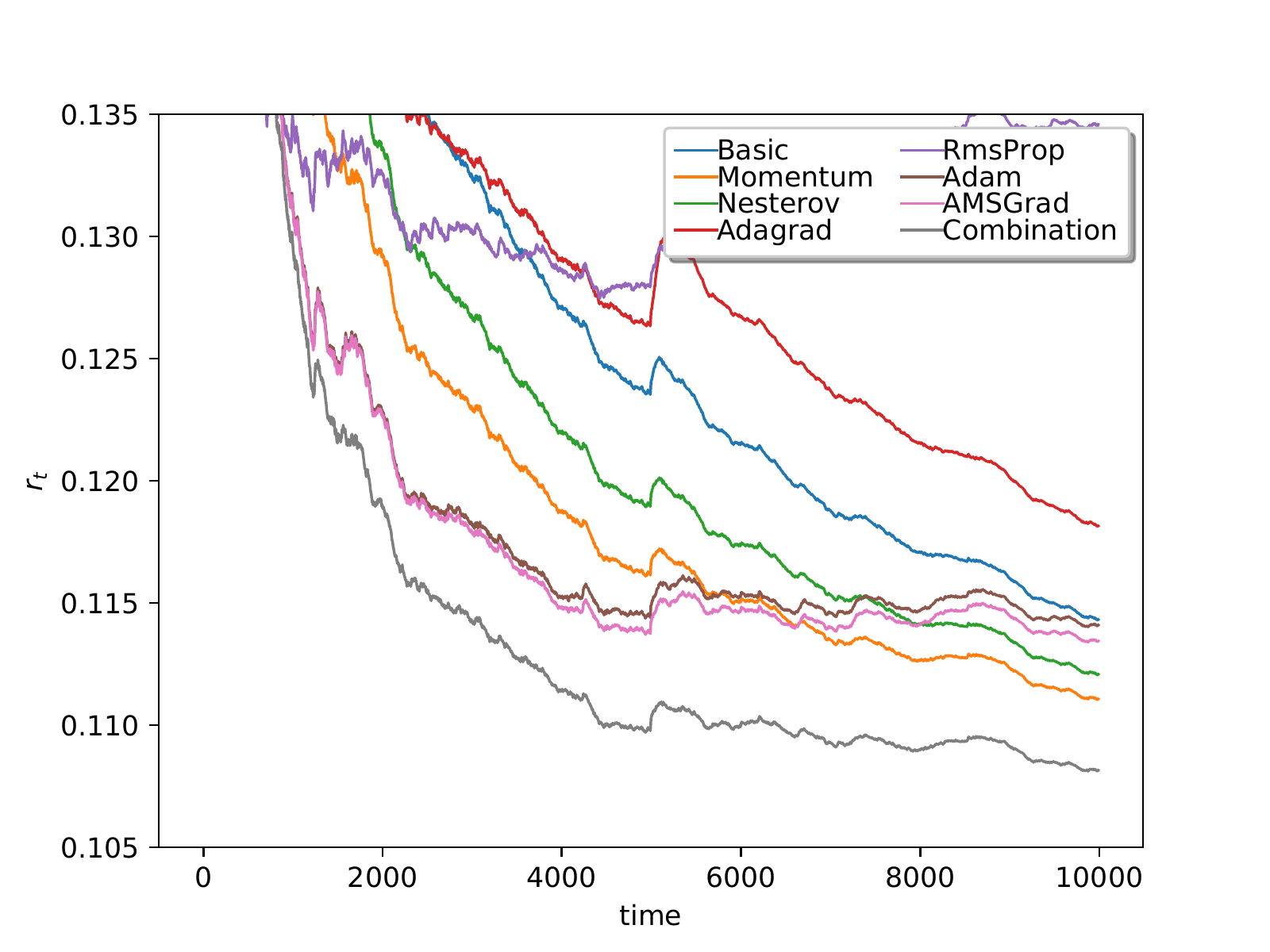}
    \caption{Average residual value for ARIMA concept shift data.}
    \label{fig:arima_conceptshift}
\end{figure}
In order to compare the quality and convergence of the different optimizers, the results of the synthetic data analysis can be seen in \Cref{fig:arima_5_0,fig:arima_5_2,fig:arima_conceptshift}. Here, no minibatches were used, instead the results are shown for every single sample. For all three settings, the grid search came to the same resulting learning rate of ${5\cdot10^{-2}}$ and hyperparameter setting of $\lambda=2000$, representing 20\% of the data. The Online ARIMA models were initialized with random coefficients in the range of [-0.5, 0.5] and the experiments were repeated 30 times, resulting in the averaged residual values $r_t$ shown in the figures.

The results for the first setting can be seen in Figure \ref{fig:arima_5_0}. Here, an Onlne ARIMA model with window size $mk=5$ and $d=0$ was trained in order to visualize the convergence of the different optimizers. It can be seen that all of the optimizers converge towards a lower residual value, while the proposed method converges fastest and best.

Figure \ref{fig:arima_5_2} shows the average residual value $r_t$ for the second setting where noise was added. Here, the window size of the Online ARIMA model was increased to $mk=10$ in order to estimate the moving average terms of the underlying ARIMA model, as proposed by \cite{liu2016online}. The results are very similar to the first setting, as the proposed approach outperforms the existing approaches. All optimizers were able to handle the added noise.

The results of the third setting from the synthetic data can be seen in Figure \ref{fig:arima_conceptshift}. The first half of the results is similar to the results of the second setting while different behavior in response to the concept shift can be observed. The RmsProp method seems to be unable to respond well to the concept shift while the other methods adapt to the new coefficients over time. The proposed approach is again able to converge fastest to the new coefficients.

\begin{figure}[t]
\centering
    \includegraphics[width=3.05in]{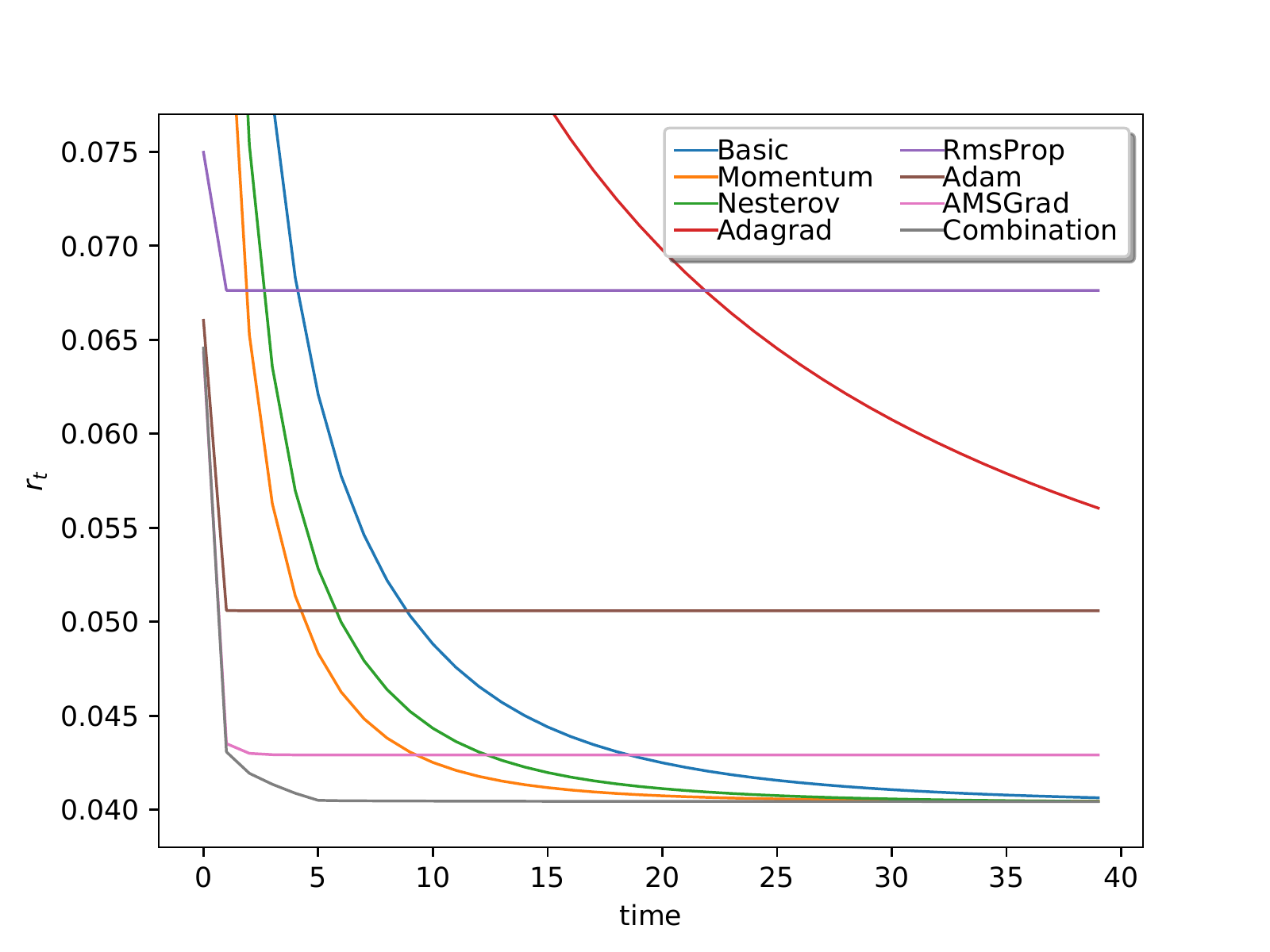}
    \caption{Average residual value during training of 1 single batch for 40 iterations with different optimizers.}
    \label{fig:nasa_1batch_dist}
\end{figure}
\begin{figure}[t]
\centering
    \includegraphics[width=3.05in]{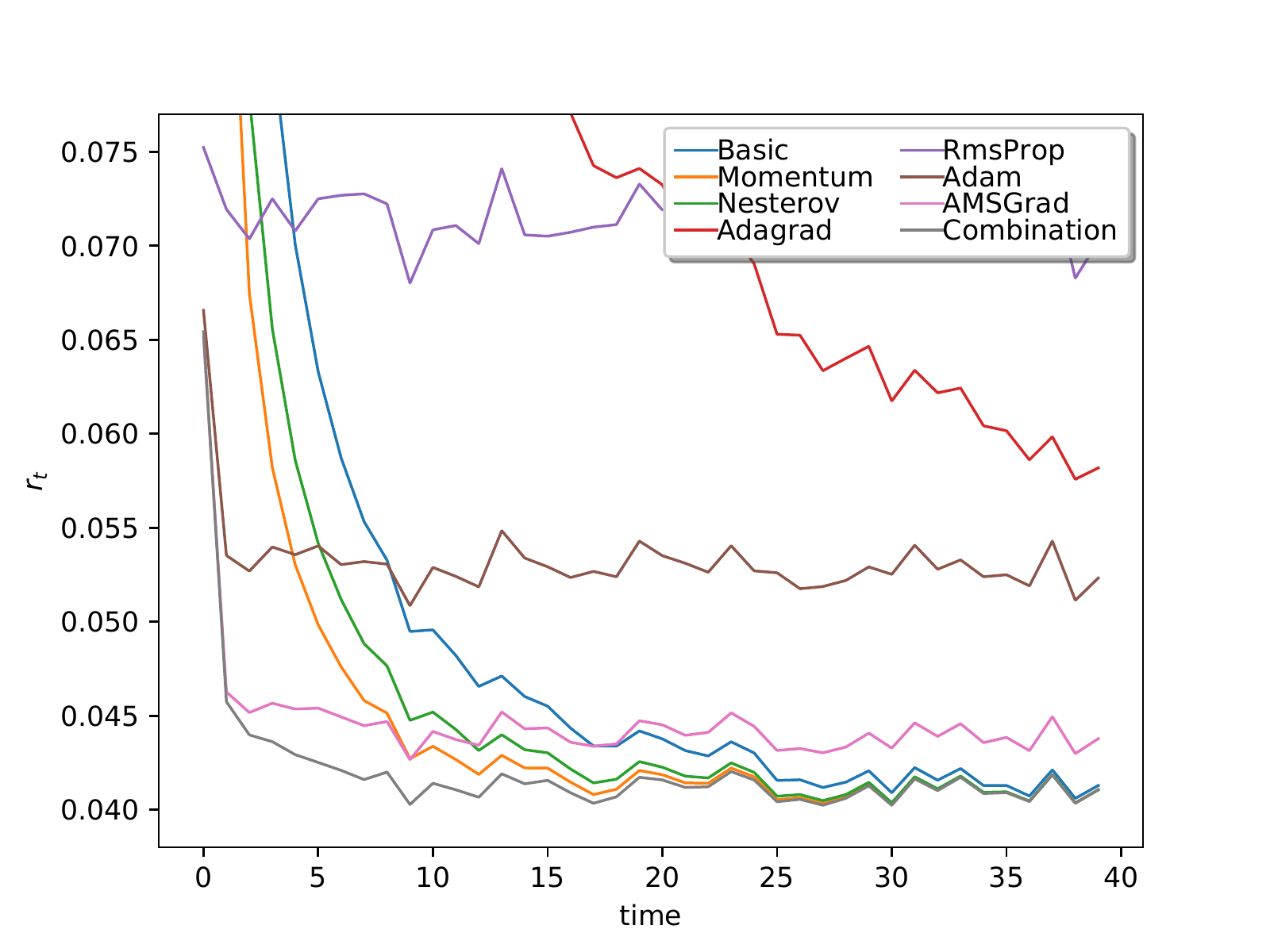}
    \caption{Average residual during training of 50 consecutive batches from the bearing dataset with different optimizers.}
    \label{fig:nasa_100batches_dist}
\end{figure}
\begin{figure}[t]
\centering
    \includegraphics[width=3.05in]{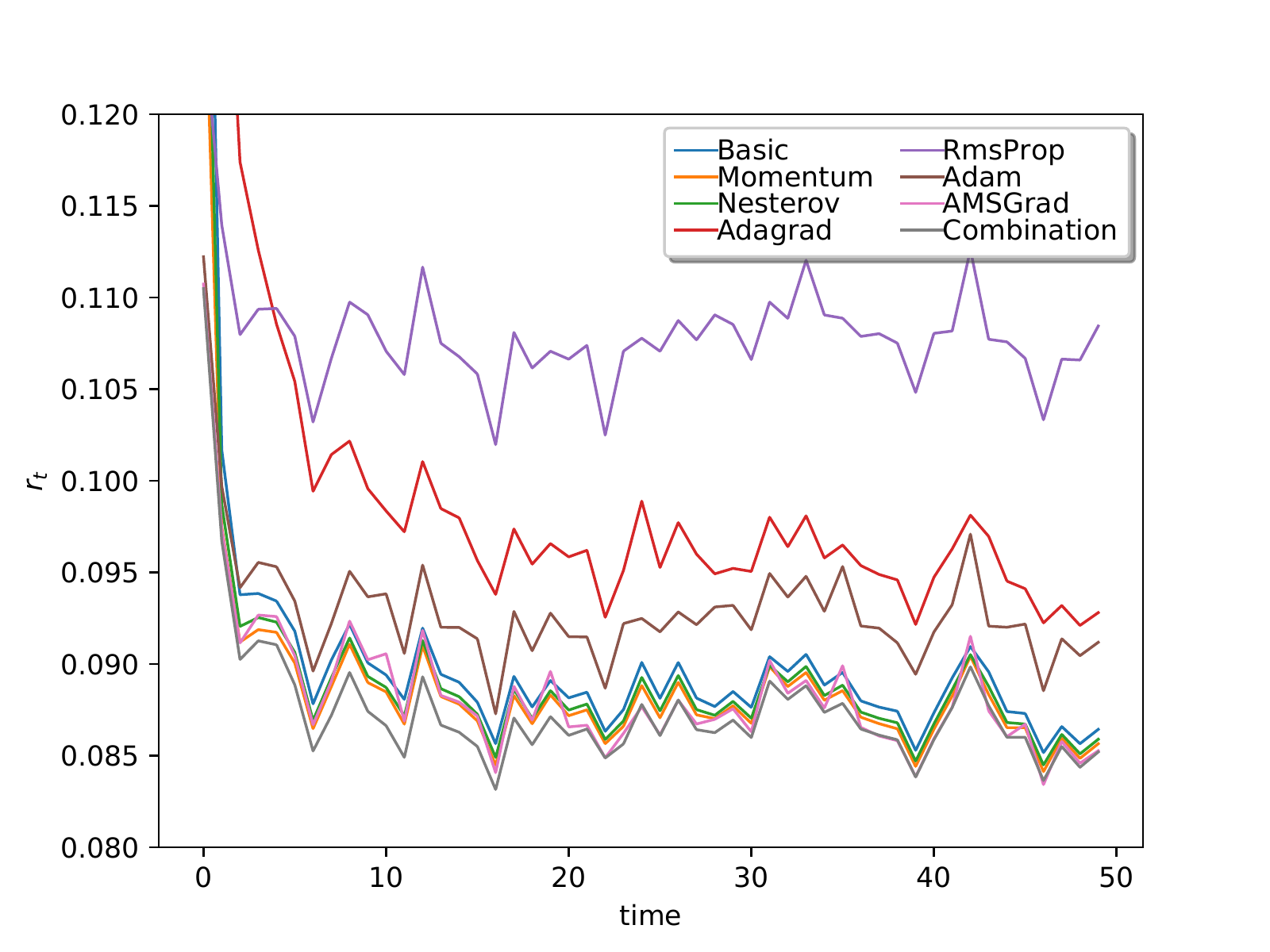}
    \caption{Average residual value during training of 50 consecutive batches from the industry dataset with different optimizers.}
    \label{fig:poc_100batches_dist}
\end{figure}

The first evaluation setup for the real-world data investigates the convergence ability of the different approaches on real-world data. Therefore, a single batch of data is learned for several iterations in order to show that Online ARIMA in conjunction with the different optimizers is able to represent and adapt to the inherent function of the time series. The results can be seen in Figure \ref{fig:nasa_1batch_dist}. Here, an Online ARIMA model with window size $mk=300$ and $d=0$ was trained with a fixed learning rate of $5\cdot10^{-3}$ and $\lambda=102400$, representing 12.5\% of the data. For the bearing dataset, the Online ARIMA models were again initialized with random coefficients in the range of $[-0.5, 0.5]$ and the experiments were repeated 10 times, results showing the averaged values.

In Figure \ref{fig:nasa_1batch_dist}, it can be observed that the optimizers Adam, RmsProp, Adagrad and AMSGrad fail to find the Online ARIMA model with the best quality as their average distance value converges to higher numbers than the other methods.
The simpler methods like the Basic, Momentum and Nesterov optimizers seem to converge towards the same average distance with different speeds. As expected, the RmsProp and AMSGrad optimizers, as well as the Adam optimizer for the first batches, seem to converge faster towards a near-optimal solution than the simpler methods. The proposed approach outperforms all other methods in terms of average distance compared over all batches.

The results of a more sophisticated experiment, where 40 consecutive measured batches of the bearing dataset have been investigated, can be seen in Figure \ref{fig:nasa_100batches_dist}. The same model configuration as for the first experiment has been used. Figure \ref{fig:nasa_100batches_dist} shows how the different optimizers behave over time in terms of quality. While Adagrad, Rmsprop, Adam are not able to achieve a quality as high as the other approaches, it can be observed that the AMSGrad optimizer is able to adapt fastest to a better model. The proposed approach again outperforms the other approaches. It can be observed that over a longer period of time the basic and momentum-based optimizers adapt to a very similar average distance as the proposed approach, while Momentum being the fastest, followed by Nesterov and finally the basic optimizer.

The same experiment was executed on the industry dataset. The results can be seen in Figure \ref{fig:poc_100batches_dist}. Here, a smaller Online ARIMA model with window size $mk=60$ and $d=1$ was used as the sample rate of the data was smaller. The learning rate was set to $1\cdot10^{-2}$ and $\lambda=102400$, representing 25\% of the data. For this dataset, the Online ARIMA models were again initialized with random coefficients in the range of $[-0.5, 0.5]$ and the experiments were repeated 10 times, results showing the averaged values.

The results look similar to the bearing dataset. Again, Adam, RMSProp and Adagrad are not able to achieve quality results similar to the other optimizers as can be seen in Figure \ref{fig:poc_100batches_dist}.
The best results are yielded by the proposed approach which is able to converge fastest to the best solution. The AMSGrad, Momentum, Nesterov and basic optimizers converge to nearly the same results the longer the experiment ran.

Overall, we showed that the proposed approach performs best in the synthetic as well as in the real world scenario compared with the stated optimizers. The proposed method was able to efficiently adapt the coefficients of an Online ARIMA model to synthetic ARIMA data as well as measured sensor data. During the beginning of the training as well as in later stages the proposed approach was able to achieve the lowest residual values.

\subsection{Hyperparameter Evaluation}
\begin{figure}[t]
\centering
    \includegraphics[width=3.05in]{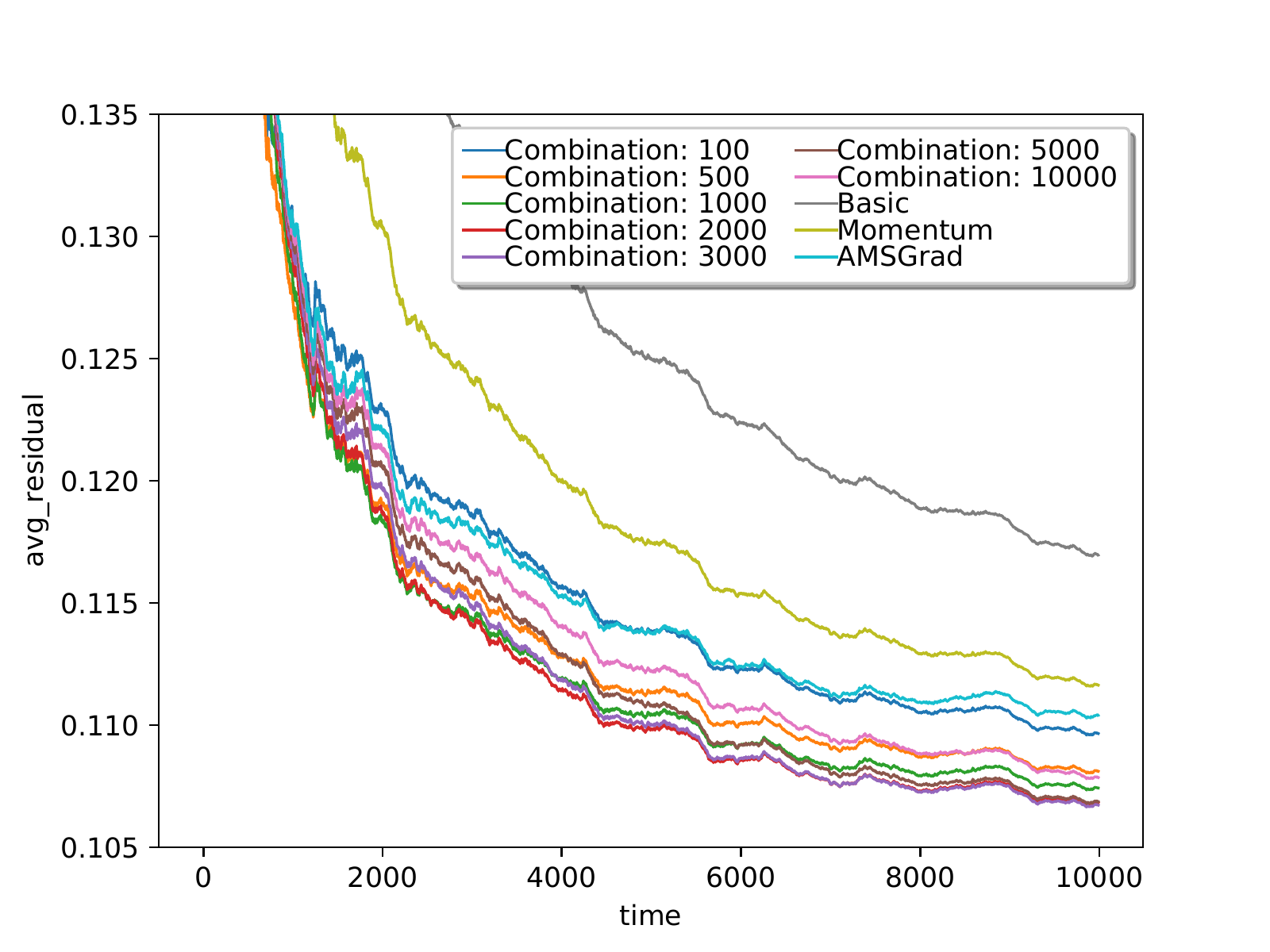}
    \caption{Average residual value for different hyperparameter settings compared to standard methods.}
    \label{fig:hyperparam2}
\end{figure}

As our approach introduced the hyperparameter $\lambda$, we conducted an evaluation showing the insensitivity of this parameter. Therefore, the second setting of the synthetic data was used with different configurations of the hyperparameters in conjunction with the best-performing standard method AMSGrad, as well as the basic and the momentum optimizer for comparability. The different examined hyperparameters $\lambda$ have been chosen from a logarithmic scale, representing a certain percentage of all training steps: 100 (=1\%), 500 (=5\%), 1000 (=10\%), 2000 (=20\%), 3000 (=30\%), 5000 (=50\%), 10000 (=100\%).

In Figure \ref{fig:hyperparam2} it can be seen that all hyperparameter configurations achieved a better final residual compared to the best performing existing method AMSGrad, although edge configurations like the 1\% and 100\% configurations were included, indicating an insensitivity of the proposed approach to the introduced hyperparameter. The overall best performing hyperparameter configuration was $\lambda=2000$ while the 3000 and 5000 setting also arrived at the same final result, indicating a broad range of viable hyperparameter configurations.

\section{Conclusion}
We proposed a novel approach combining OGD optimizers during the training by gradually changing the used optimizer during the training. We provided a function to calculate the linear combination and the experiments have shown that the proposed approach outperforms the existing approaches. In all conducted experiments, the overall error during forecasting was lower than the error of the compared methods at any point of training. Thus, the proposed approach was able to adapt to the underlying process fast and afterwards fine-tune the coefficients of the model in order to increase the quality.

As future work, we will focus on the combination of the proposed approach with learning rate schedules as well as an investigation of the transferability onto the learning of neural networks.

\bibliography{bibliography}
\bibliographystyle{IEEEtran}

\end{document}